\documentclass{sig-alternate-05-2015}
\usepackage{graphicx}
\usepackage{multirow}
\usepackage{subfigure}

\begin{document}

% Copyright
%\setcopyright{acmcopyright}
%\setcopyright{acmlicensed}
%\setcopyright{rightsretained}
%\setcopyright{usgov}
%\setcopyright{usgovmixed}
%\setcopyright{cagov}
%\setcopyright{cagovmixed}

% DOI
%\doi{10.475/123_4}

% ISBN
%\isbn{123-4567-24-567/08/06}

%Conference
%\conferenceinfo{PLDI '13}{June 16--19, 2013, Seattle, WA, USA}

%\acmPrice{\$15.00}

%
% --- Author Metadata here ---
%\conferenceinfo{WOODSTOCK}{'97 El Paso, Texas USA}
%\CopyrightYear{2007} % Allows default copyright year (20XX) to be over-ridden - IF NEED BE.
%\crdata{0-12345-67-8/90/01}  % Allows default copyright data (0-89791-88-6/97/05) to be over-ridden - IF NEED BE.
% --- End of Author Metadata ---

\title{NIST: An Image Classification Network to Image Semantic Retrieval
%\titlenote{(Produces the permission block, and
%copyright information). For use with
%SIG-ALTERNATE.CLS. Supported by ACM.}}
}

\numberofauthors{3} %  in this sample file, there are a *total*
\author{
% 1st. author
\alignauthor
Xiuyuan Chen\\
       \affaddr{Univ. of Electronic Science and Technology of China}\\
       \email{fqhshmily@163.com}
% 2nd. author
\alignauthor
Le Dong\\
       \affaddr{Univ. of Electronic Science
and Technology of China}\\
       \email{ledong@uestc.edu.cn}
% 3rd. author
\alignauthor
Mengdie Mao\\
       \affaddr{Univ. of Electronic Science
and Technology of China}\\
       \email{maomengdie0910@163.com}
 % 4rd. author
%\additional
%Qianni Zhang\\
 %      \affaddr{Queen Mary Univ. of London}\\
%       \email{qianni.zhang@qmul.ac.uk}
}
% There's nothing stopping you putting the seventh, eighth, etc.
% author on the opening page (as the 'third row') but we ask,
% for aesthetic reasons that you place these 'additional authors'
% in the \additional authors block, viz.

% Just remember to make sure that the TOTAL number of authors
% is the number that will appear on the first page PLUS the
% number that will appear in the \additionalauthors section.

\maketitle
\begin{abstract}
This paper proposes a classification network to image
semantic retrieval (NIST) framework to counter the image retrieval challenge.
Our approach leverages the successful classification network
GoogleNet based on Convolutional Neural Networks to
obtain the semantic feature matrix which contains the serial number of
classes and corresponding probabilities. Compared with
traditional image retrieval using feature matching to
compute the similarity between two images, NIST
leverages the semantic information to construct semantic
feature matrix and uses the semantic distance algorithm
to compute the similarity. Besides, the fusion
strategy can significantly reduce storage and time consumption
due to less classes participating in the last semantic
distance computation. Experiments demonstrate that our NIST framework produces state-of-the-art results
in retrieval experiments on MIRFLICKR-25K dataset.

\end{abstract}





%
%  Use this command to print the description
%

% We no longer use \terms command

\section{Introduction}
Image retrieval has long been a challenging task in
computer vision, especially when the image amount
gets continually increasing. The semantic gap between
the low-level visual features and high-level semantics \cite{datta2008image}
and the intention gap between user's search intent
and the query \cite{hanjalic2012intent, Zha2009Visual} have long been a big challenge. Numerous
efforts have been made to counter this significant
challenge, among which the convolutional neural
network (CNN) has recently demonstrated impressive
progress. Due to the extensive use of CNN, a lot of
networks based on CNN obtained extremely high accuracy.
Therefore, we leverage a successful classification
network called GoogleNet \cite{szegedy2015going} to generate semantic
information of input images. For us, excellent classification
results is the first step of the accurate image retrieval
due to the exact semantic information has significant
influence on feature representation.

\begin{figure}
\begin{center}
\includegraphics[width=1\linewidth]{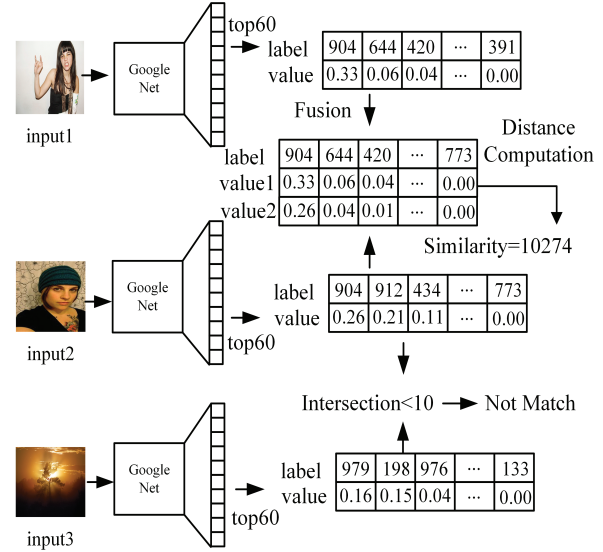}
\end{center}
\caption{NIST framework.}
\label{fig1}
\end{figure}
In this paper, we introduce a classification network to image
semantic retrieval (NIST) framework based on deep learning to learn the
probability of classification results, fuse the semantic
feature matrix and use the matrix to compute the
similarity between multi-label images in the
semantic space. Figure 1 illustrates the NIST framework
of three parts. Particularly,
given a query image, our goal is to use several representative
dimensions and the corresponding probabilities outputing from
GoogleNet to represent the semantic information. Note that each
semantic concept is viewed as a class, one or more classifiers
are trained from the training data. Meanwhile, we use the
probability of class to evaluate the integrating degree
between input image and the class of the current dimension.
After generating these structured outputs from the
classification network, we leverage these results to
construct a semantic feature matrix contains the serial
number and corresponding probability of each class. When
comparing two images, we firstly choose $K$ classes
with the top $K$ highest probabilities to be a new matrix
and then fuse the two matrixes by our fusion strategy to
obtain a 3-dimensional matrix. Finally, we compute the similarity
of two images by the semantic distance computation algorithm.

The contributions of this paper can be concluded as
follows: (1) offering a train of thoughts from a convolutional
neural network which efficiently process classification
tasks to image semantic retrieval, designing a semantic
distance computation algorithm according to semantic
feature extracted from CNN; (2) proposing an efficient
semantic feature. The feature matrix only uses less than 100 features
to represent the semantic information of images, having
significant influence on storage saving, time consumption and
match efficiency. This new feature matrix can fuse with
other image features to improve the retrieval accuracy.
\section{Our Approach}
In general, a semantic feature matrix $S_2*n$ is treated
as a mapping that represents an input onto a matrix
including the serial number of each class and corresponding
probabilities. Assume that we are given a set of class labels
$L = \left\{1, \cdots, N\right\}$, our goal is to learn a set
of probabilities $F_{1} =\left\{f_{1},f_{2}, \cdots ,f_{i}, \cdots ,f_{n}\right\}$
of each class that generates semantic feature matrix $S_2*n$.

\subsection{Classification Network}
In recent years, the quality of image classification has been progressing at a dramatic pace mainly due to the advances of deep learning, more concretely convolutional networks \cite{lecun1989backpropagation}.
Among which the GoogleNet obtains a top-5 error of 6.67\% on
both the validation and testing data, ranking the first among
other participants in the ImageNet Large-Scale Visual Recognition
Challenge 2014 (ILSVRC14). In our work, we leverage the
GoogleNet to train our data and obtain the classification results.

We leverage the classification results obtained from GoogleNet
containing the serial number of each class ranking from 1 to
1000, and the corresponding probabilities denoted as $F_{1} =\left\{f_{1},f_{2}, \cdots ,f_{i}, \cdots ,f_{n}\right\}$
to represent the semantic information of input image.
Note that each serial number of class has a
probability $f_i$ to represent the similar
degree between input image and the current class.
Giving each class an unique serial number and using the serial
number sequence and corresponding probabilities to construct
a 2-dimensional matrix $S_2*n$.

\subsection{Semantic Feature Extraction and Fusion}
The classification results have shown the semantic information
of input images, whereas some low possibility classes are insignificant
to semantic representation. Chose some highest probability classes
left can not only save storage, but also reduce the computation
time of semantic distance. By experimental validation illustrated
in Figure 2, we find that choose 60 classes may have the most desirable
experiment results. So we rank the probabilities of 1000
classes, and extract top 60 highest probability classes and
its corresponding probabilities to form a new matrix which
compactly denotes the semantic information of images.
\begin{table}
 \centering
 \caption{Semantic feature matrix of input images.}
  \begin{tabular}{|p{3.8cm}|p{0.2cm}|p{0.2cm}|p{0.2cm}|p{0.2cm}|p{0.2cm}|p{0.6cm}|}
    \hline
       \textbf{Serial number of classes} & 1 & 2 & ... & i & ... & 1000\\\hline
       \textbf{Probability} & $f_1$ & $f_2$ & ... & $f_i$ & ... & $f_{1000}$\\\hline
  \end{tabular}
\end{table}
When comparing two images, we firstly merge the two
compact semantic feature matrix to form a new 3-dimensional
matrix. First, we compare the serial number of classes of
two images. If the number of the same serial number
between two images is less than 10, we may think these two
images have low similarity thus do not need to merge and
compute the similarity between them. After the first coarse
filter, the process of merging the two matrix is a recursion
followed by the following steps: (1) If they have the same serial
number, just merge the serial number and write the
corresponding probability in the second and third row in the
matrix, respectively. (2) If the serial number is not the same,
write the serial number in a new column vector
with the probability in the corresponding row for
which image has this serial number. And write the probability
with 0 in the row for which image does not have this serial
number of class. (3) For each serial number, repeat the second and third step.

\subsection{Semantic Distance Computation}
After the process of fusing two compared images, we get
a $S_3*n$ matrix containing the serial number and corresponding
probabilities of these 60 classes. Our fusion strategy may bring
some probability with 0, this will cause negative influence on
semantic distance computation. Considering some extreme cases,
we add parameter $M_1$ to increase the positive influence
of semantic distance, and add parameter $M_2$ to
alleviate the negative influence at the same time. For instance, if two images
have a lot of identical classes, meanwhile the probability
of each class is relative high as the same as another one,
these two images can get a relatively high semantic distance.
Or the probability of each class is different with another
one to a large extent, the semantic distance between two images
may affect by negative influences.

Particularly, we use datum in the $S_3*n$ matrix to compute the
semantic distance between two images by the following distance algorithm:
\begin{displaymath}\
D = \frac{M_1{\sum_{i=0}^{i=K}f_{1i}f_{2i}} - M_2{\sum_{i=0}^{i=K}(f_{1i} - f_{2i})^2}}{max(f_{1i}f_{2i})}.
\end{displaymath}
where $K$ is obtained by counting the number of classes in
the $S_3*n$ matrix, ranging from 60 to 120. $f_{1i}$ and $f_{2i}$ represent the probability of
two images in the same serial number of class, respectively. Weight $M_1$ and $M_2$
are got by experimental validation. $max(f_{1i}f_{2i})$ is used to normalize the
similarity and alleviate the big gap existing in the probabilities
of the same class of two images.

\section{Experimental Evaluation}
We evaluate the performance of NIST framework on
the multi-label benchmark dataset MIRFLICKR-25K \cite{huiskes2008mir}. We present quantitative evaluations in terms
of ranking measures and compare NIST with unsupervised
methods iterative quantization (ITQ) \cite{gong2013iterative} and
spectral hashing (SH) \cite{weiss2009spectral}, and supervised
methods  CCA-ITQ \cite{gong2013iterative}, hamming distance metric learning (HDML) \cite{norouzi2012hamming}
using multi-label and ranking information, respectively. Normalized Discounted
Cumulative Gain (NDCG) \cite{jarvelin2000ir} and Average
Cumulative Gain (ACG)  \cite{jarvelin2000ir} are used as metrics
to evaluate our NIST framework. More particular experimental and parameter settings are
presented in subsequent sections.

\subsection{Dataset}
The MIRFLICKR-25K dataset consists of 25,000 images
collected from the social photography site Flickr.com through
the public API. For each image, it has a description tag text
file, a camera information file and a copyright license file.
All images are annotated for 24 semantic concepts including
various scenes and objects categories such as sky, water,
beach and dog. All 25,000 images are used as the database
for testing queries and retrieval due to the very successful
GoogleNet has been well trained. Following \cite{szegedy2015going}, a
1000-dimensional feature vector for each image is extracted
by GoogleNet, which will be used for the process of feature
fusion and compared methods.

\subsection{Evaluation Criteria}
Compared with the deep semantic ranking
based hashing (DSRH) framework \cite{zhao2015deep} using deep
convolutional neural network to construct
hash functions to learn directly from images, NIST use CNN to
obtain classification results and propose the semantic distance
algorithm to compute the similarity of two
images. In our experiments, NDCG and ACG are used to measure the
ranking quality of retrieved database points in order to
compare with these image semantic retrieval methods based on hash function.

NDCG is a popular measurement in the information retrieval community,
it evaluates the ranking of data points by penalizing errors in higher ranked
items more strongly. NDCG is defined as:
\begin{displaymath}\
NDCG@p = \frac{1}{Z}\sum_{i=1}^{p}\frac{2^{r_i}-1}{\log(1+i)}.
\end{displaymath}
where $p$ is the truncated position in a ranking list, $Z$ is a normalization
constant to ensure that the NDCG score for the correct
ranking is one, and $r_i$ is the similarity level of the $i-th$ database
point in the ranking list.

ACG is calculated by taking the average of the similarity
levels of data points within top-$p$ positions and it is equivalent to the
precision weighted by the similarity level of each data point. ACG is defined as:
\begin{displaymath}\
ACG@p = \frac{1}{p}\sum_{n=1}^{p}{r_i}.
\end{displaymath}
where $p$ is also the truncated position in a ranking
list, $r_i$ is the similarity level of the $i-th$ database
point in the ranking list. In our experiment, we set $p=100$.

\subsection{The Effect of Parameters}
We discuss the parameter $K$ which means the truncated
position in the ranking list of 1000-dimensional feature
vector for each image extracted by GoogleNet. We set
$K =\left\{20,30,40,50,60\right\}$, considering the biger $K$ we set,
the more storage and computation consumption we may cost. Figure 2
shows the results of the influence of
different $K$  on retrieval accuracy in the ranking measurements.
We can observe that with the increase of $K$, the retrieval accuracy increases
at the same time. It's obvious that NIST obtains the best
performance at $K=60$, and the accuracy increases are tiny
when $K$ range from 50 to 60 both in NDCG
and ACG. This is because of that the probabilities in semantic feature
matrix which smaller than the top 60 is very small, having less
significant in representing the semantic information of images.
According to this, we just set $K=60$ in the following experiments.

%\begin{figure}[t]
%\centering
%\includegraphics[width=1.7\linewidth]{kkk.eps}
%\scalebox{1}[1]{\includegraphics[width=1\linewidth]{kkk.pdf}}
%\caption{The influence of different K.}
%\label{fig1}
%\end{figure}

%\begin{table}
% \centering
 %\caption{The influence of different K.}
%  \begin{tabular}{|p{2cm}|p{0.9cm}|p{0.9cm}|p{0.9cm}|p{0.9cm}|p{0.9cm}|}
%    \hline
 %      \textbf{$K$} & 20 & 30 & 40 & 50 & 60\\\hline
 %      \textbf{NDCG@100} & 39.74\% & 43.51\% & 45.83\% & 46.76\% & 46.92\%\\\hline
 %      \textbf{ACG@100} & 2.53 & 2.65 & 2.70 & 2.83 & 2.87\\\hline
 % \end{tabular}
%\end{table}

To analyze the relevant between performance of NIST
and the value of $M_1$/$M_2$, we validate the
influence of different values on retrieval accuracy in the
experiments. The results in the ranking measurements are
showed in Figure 3. As mentioned before, $M_1$ and $M_2$ are
used to increase the positive influence and reduce the negative
influence respectively, so we think $M_1$ should much bigger
than $M_2$. So we set the $M_1/M_2 = \left\{2000,5000,10000,50000\right\}$. From the
experimental results, we can observe that with the increase
of the value of $M_1/M_2$ raging from 2000 to 10000, a big rising
of retrieval accuracy is apparent. But from 10000 to 50000, there is a
little rise of accuracy when using ACG as the measurement. However, there is a
decrease of accuracy when using NDCG as the measurement. Therefore, when
the value of $M_1$/$M_2$ is 10000, we can get the best performance.
So we just set $M_1$/$M_2$=10000 in the following experiment.
%\begin{table}
 %\centering
% \caption{The influence of different $M_1$/$M_2$.}
%  \begin{tabular}{|p{2cm}|p{0.9cm}|p{0.9cm}|p{0.9cm}|p{0.9cm}|}
%    \hline
 %      \textbf{$M_1/M_2$} & 2000 & 5000 & 10000 & 50000\\\hline
 %      \textbf{NDCG@100} & 30.32\% & 43.45\% & 46.92\% & 46.65\%\\\hline
 %      \textbf{ACG@100} & 1.82 & 2.37 & 2.87 & 2.90\\\hline
 % \end{tabular}
%\end{table}

\begin{figure}
\begin{center}
\includegraphics[width=1\linewidth]{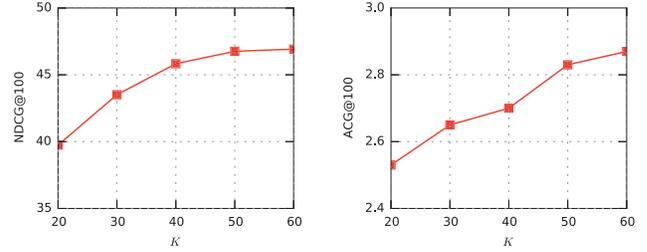}
\end{center}
\caption{The effect of $K$.}
\label{fig2}
\end{figure}

\begin{figure}
\begin{center}
\includegraphics[width=0.49\linewidth]{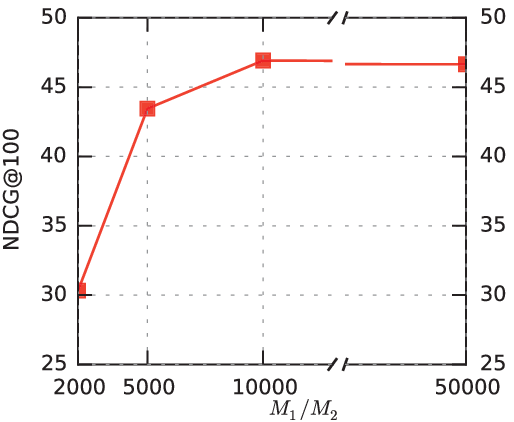}
\includegraphics[width=0.49\linewidth]{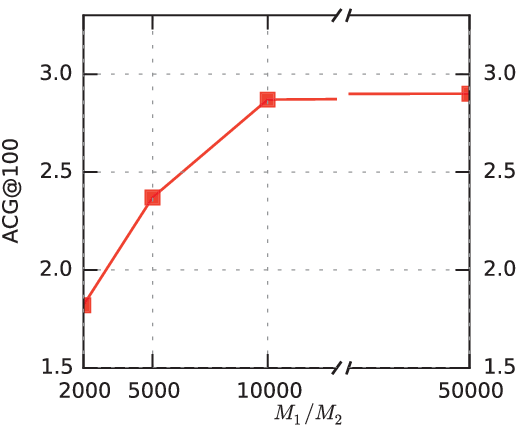}
\end{center}
\caption{The effect of $M_1$/$M_2$.}
\label{fig3}
\end{figure}

\subsection{Comparison}
We also compare the proposed NIST framework with other hash
methods based on hand-crafted features and
DSRH framework which based on CNN. Table 2 illustrates
the scores of these methods using NDCG and ACG as metrics. We can
observe that the performance of NIST is better
than other methods based on hand-crafted features
in all cases. However, it's obvious that NIST has some gaps
between DSRH framework in retrieval accuracy.

\begin{table*}
 \centering
 \caption{Comparison of retrieval accuracy of different methods on MIRFLICKR-25K.}
  \begin{tabular}{|p{2cm}|p{1.4cm}|p{2.7cm}|p{2.8cm}|p{2.4cm}|p{2.2cm}|}
    \hline
       \textbf{Method} & Ours & DSRH-64bits \cite{zhao2015deep} & CCA-ITQ-64bits \cite{gong2013iterative} & HDML-64bits \cite{norouzi2012hamming} & SH-64bits \cite{weiss2009spectral}\\\hline
       \textbf{NDCG@100} & 46.92\% & 50.41\% & 31.15\% & 22.03\% & 18.54\%\\\hline
       \textbf{ACG@100} & 2.87 & 3.16 & 2.3 & 1.75 & 1.46\\\hline
  \end{tabular}
\end{table*}

There are some important factors causing the gap between
NIST and the DSRH. Due to some sematic tags used in the test set such
as night, transport, plant life have not trained
in the training set, and we have not retraining the
network by using the MIRFLICKR-25K dataset. The
classification network GoogleNet we chose put more emphasis
on single-label classification tasks, having
more disadvantages on multi-label classification tasks. According
to this, we can guess that a multi-label
classification network may bring us much better accuracy. However,
experiments show that it's reasonable for our novel NIST framework using
classification results to obtain sematic information of compared images.

\subsection{Discussion}
In the process of using GoogleNet to classify the images from MIRFLICKR-25K dataset,
we find that single label images can get a more accurate classification
result getting from GoogleNet. However, multi-label images' classification
results are not accurate enough. Particularly, NIST is
relatively depending on the classification network.
We choose 100 images of the worst retrieval results obtained by NDCG
measurement as input images, when using GoogleNet to classify these
images, the results show that there are only 3 images' classification
results are reasonable, the classification results of other 97 images
having a big gap from the groundtruth. If we only choose the 100 best
retrieval results as input images and use GoogleNet to classify,
the retrieval accuracy may larger than 85\% by NDCG measurement.
This increases in accuracy obviously demonstrate that NIST is reasonable, but
it needs a more accurate multi-label classification network.
In future work, we may train a classification network having
advantages in both single label and multi-label images.

\begin{figure}
\begin{center}
\includegraphics[width=1\linewidth]{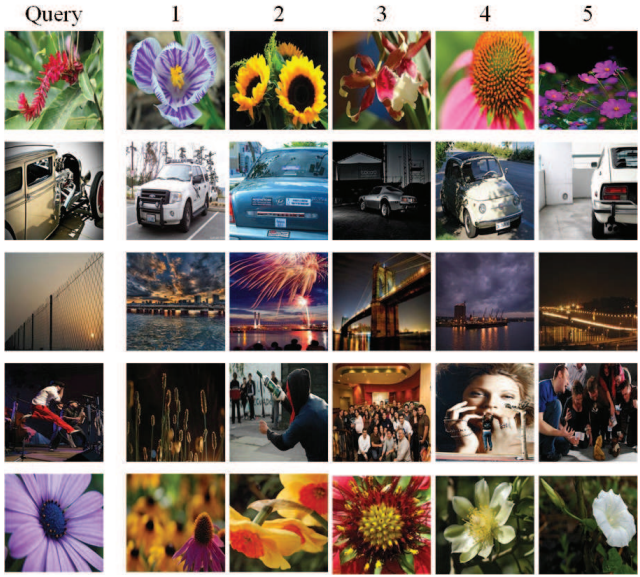}
\end{center}
\caption{Three sample queries from the MIRFLICKR-25K dataset and their corresponding retrieval results.}
\label{fig1}
\end{figure}

%\begin{table*}
%\centering
%\caption{Some Typical Commands}
%\begin{tabular}{|c|c|l|} \hline
%Command&A Number&Comments\\ \hline
%\texttt{{\char'134}alignauthor} & 100& Author alignment\\ \hline
%\texttt{{\char'134}numberofauthors}& 200& Author enumeration\\ \hline
%\texttt{{\char'134}table}& 300 & For tables\\ \hline
%\texttt{{\char'134}table*}& 400& For wider tables\\ \hline\end{tabular}
%\end{table*}
% end the environment with {table*}, NOTE not {table}!

%\begin{figure}
%\centering
%\scalebox{1.2}[1.2]{\includegraphics[width=0.8\linewidth]{kkk.pdf}}
%\caption{A sample black and white graphic.}
%\end{figure}

%\begin{figure*}
%\centering
%\includegraphics{flies}
%\caption{A sample black and white graphic
%that needs to span two columns of text.}
%\end{figure*}

%\begin{figure}
%\centering
%\includegraphics[height=1in, width=1in]{rosette}
%\caption{A sample black and white graphic that has
%been resized with the \texttt{includegraphics} command.}
%\vskip -6pt
%\end{figure}

\section{Conclusions}
In this paper, we propose a new NIST framework for image semantic
retrieval. Pursuing the compactness without compromising
the retrieval accuracy, the fusion strategy and sematic
distance computation algorithm are designed to accomplish NIST framework
in a straightforward manner. Experiments demonstrate that
NIST framework show its reasonability and expansibility on
the semantic retrieval tasks. In future work, we would
like to use a multi-label classification network to obtain
more accurate classification results.
%\end{document}  % This is where a 'short' article might terminate

%ACKNOWLEDGMENTS are optional
%\section{Acknowledgments}
%This work was supported in part by the National Natural
%Science Foundation of China under Grant 61370149, in
%part by the Fundamental Research Funds for the Central
%Universities(ZYGX2013J083), and in part by the
%Scientific Research Foundation for the Returned
%Overseas Chinese Scholars, State Education Ministry.

\bibliographystyle{abbrv}
\bibliography{sigproc}  % sigproc.bib is the name of the Bibliography in this case
% ACM needs 'a single self-contained file'!
%\balancecolumns

\end{document}